\documentclass[10pt,twocolumn,letterpaper]{article}

\usepackage{iccv}
\usepackage{times}
\usepackage{epsfig}
\usepackage{graphicx}
\usepackage{amsmath}
\usepackage{amssymb}

\usepackage{multirow}
\usepackage{multicol}
\usepackage{lipsum}
\usepackage{color}
\usepackage{bm}

\usepackage{times}
\usepackage{microtype}
\usepackage{epsfig}
\usepackage[table,xcdraw]{xcolor}
\usepackage{caption}
\usepackage{float}
\usepackage{placeins}
\usepackage{color, colortbl}
\usepackage{stfloats}
\usepackage{enumitem}
\usepackage{tabularx}
\usepackage{xstring}
\usepackage{multirow}
\usepackage{xspace}
\usepackage{url}
\usepackage{subcaption}
\usepackage{xcolor}
\usepackage[hang,flushmargin]{footmisc}
\usepackage[table]{xcolor}
\usepackage[pagebackref, breaklinks=true, colorlinks, citecolor=citecolor, linkcolor=linkcolor, bookmarks=false]{hyperref}

\usepackage[utf8]{inputenc} 
\usepackage[T1]{fontenc}    
\usepackage{url}            
\usepackage{booktabs}       
\usepackage{amsfonts}       
\usepackage{amsmath}
\usepackage{nicefrac}       
\usepackage{microtype}      
\usepackage[table]{xcolor}  
\usepackage{tabularx}
\usepackage{graphicx}
\usepackage{multirow}
\usepackage{enumitem}
\usepackage{pifont}

\definecolor{citecolor}{HTML}{0071BC}
\definecolor{linkcolor}{HTML}{ED1C24}
\usepackage[capitalize]{cleveref}
\crefname{section}{Sec.}{Secs.}
\crefname{table}{Table}{Tables}
\crefname{figure}{Fig.}{Figs.}
% ...

\newcommand{\R}[1]{{%
    \textbf{%
        \ifstrequal{#1}{1}{\textcolor{red}{R#1}}{%
        \ifstrequal{#1}{2}{\textcolor{blue}{R#1}}{%
        \ifstrequal{#1}{3}{\textcolor{magenta}{R#1}}{%
        \ifstrequal{#1}{4}{\textcolor{teal}{R#1}}{%
                           \textcolor{cyan}{R#1}%
        }}}}%
    }%
}}

\graphicspath{{figures/}}

\usepackage[breaklinks=true,bookmarks=false]{hyperref}

\iccvfinalcopy 

\def\iccvPaperID{1869} 

\ificcvfinal\fi

\begin{document}

\title{Bridging Vision and Language Encoders:\\ Parameter-Efficient Tuning for Referring Image Segmentation}

\author{
    Zunnan Xu$^{1}$\footnotemark[1] \quad
    Zhihong Chen$^{2,3}$\footnotemark[1] \quad 
    Yong Zhang$^{4}$ \quad
    Yibing Song$^5$ \quad 
    Xiang Wan$^{3}$ \quad 
    Guanbin Li$^{1}$\footnotemark[2]\\
    $^{1}$Sun Yat-sen University \quad $^{2}$The Chinese University of Hong Kong, Shenzhen \\
    $^{3}$Shenzhen Research Institute of Big Data \quad $^{4}$Tencent AI Lab \quad $^{5}$AI$^3$ Institute, Fudan University\\
    {\tt\small xuzn3@mail2.sysu.edu.cn ~ zhihongchen@link.cuhk.edu.cn} \\
    {\tt\small \{zhangyong201303, yibingsong.cv\}@gmail.com ~ wanxiang@sribd.com ~ liguanbin@mail.sysu.edu.cn}
}

\maketitle
\ificcvfinal\thispagestyle{empty}\fi
\def\thefootnote{*}\footnotetext{Equal contribution}
\def\thefootnote{\dag}\footnotetext{Corresponding author}
\def\thefootnote{\arabic{footnote}}

\begin{abstract}
Parameter Efficient Tuning (PET) has gained attention for reducing the number of parameters while maintaining performance and providing better hardware resource savings, but few studies investigate dense prediction tasks and interaction between modalities. 
In this paper, we do an investigation of efficient tuning problems on referring image segmentation. We propose a novel adapter called Bridger to facilitate cross-modal information exchange and inject task-specific information into the pre-trained model. We also design a lightweight decoder for image segmentation. Our approach achieves comparable or superior performance with only 1.61\% to 3.38\% backbone parameter updates, evaluated on challenging benchmarks.
The code is available at \url{https://github.com/kkakkkka/ETRIS}.
\end{abstract}

\section{Introduction}
Referring image segmentation (RIS) aims to predict a mask for the target object described by a given natural language sentence based on the input image and text.
This task is distinct from semantic segmentation, which assigns each pixel in an image with a label from a fixed word set.
Instead, RIS needs to recognize the objects indicated by the language expression, which is of greater complexity due to its arbitrary context length and involving an open-world vocabulary such as object names, attributes, positions, etc. 

\begin{figure}[t]
\centering
\includegraphics[width=1.0\linewidth]{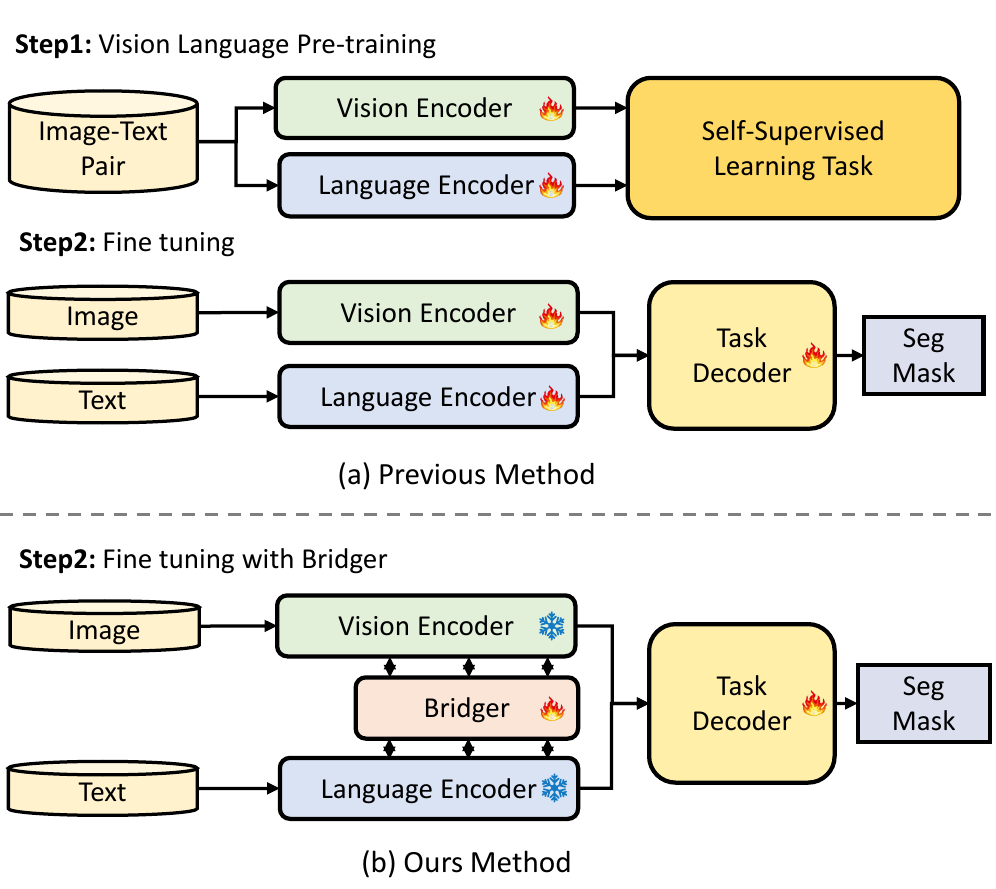}
\caption{Previous method vs. our method.
(a) The conventional method pre-trains visual language models on datasets with image-text pairs using self-supervised learning and fine-tunes them on downstream tasks.
(b) We propose Bridger, an Adapter-like module that incorporates inductive biases and task-specific information into the pre-trained model.}
\label{fig:demo}
\end{figure}

Recent studies~\cite{wang2021cris,ding2022vlt,chen2023advancing} have shown the effectiveness of fine-tuning general-purpose pre-trained models for visual grounding.
However, these approaches have a separate copy of fine-tuned model parameters for each dataset, making it expensive to deploy models across multiple scenarios. 
This issue is particularly significant for large-scale pre-trained models, which now consist of hundreds of millions to trillions of parameters~\cite{li2021grounded,zhou2022learning,chen2022pali}.

Therefore, we ask an essential question: \textit{can the model maintain a competitive performance with pre-trained backbone network parameters fixed?}.
Various parameter-efficient training methods~\cite{guo2020parameter,houlsby2019parameter,gao2021clip,chen2022adaptformer,chen2022vision,zhou2022learning} have been proposed to achieve a balance between parameter efficiency and performance.
However, most of the existing methods are limited to either single-modal tasks~\cite{guo2020parameter,houlsby2019parameter,chen2022vision} or simple classification tasks~\cite{gao2021clip,chen2022adaptformer,zhou2022learning} with few studies focusing on dense prediction tasks and the interaction between different modalities, which limits their generality.

We aim to adapt pre-trained vision-language models for referring image segmentation with comparable performance to full fine-tuning, but in a more parameter-efficient way, as demonstrated in Figure~\ref{fig:demo}.
This approach improves adaptability and eliminates the parameter inefficiencies and prohibitive expenses associated with previous methods that require creating separate copies of fine-tuned backbone model parameters for each dataset.
In detail, firstly, we introduce an additional network named Bridger that does not require pre-training and can be seamlessly integrated into the original architecture of the pre-trained model, where we introduce vision-specific inductive biases and facilitate interaction between the dual encoder.
There are two tailored modules for Bridger:
(i) a spatial prior module for capturing the local semantics (spatial prior) from feature maps of the intermediate layer and (ii) a cross-modal attention module that enables information exchange between the two modalities. 
Secondly, we designed a \textit{lightweight} task-specific decoder for referring image segmentation to make further alignment on visual and linguistic features.
Under this framework, the backbone network can be any general-propose (dual-encoder) model that is pre-trained on vision-language datasets, and we adopt CLIP~\cite{radford2021learning}, a pre-trained image-text alignment model, as our vision and language encoders.
As a result, utilizing ViT~\cite{dosovitskiy2020image} and ResNet~\cite{he2016deep} as the visual backbone and updating only 1.61\% to 3.38\% parameters, our framework achieves comparable or even better performance than previous methods which employ the same backbone for full-finetuning.
Our main contributions are as follows:
\begin{itemize}[leftmargin=*,noitemsep,nolistsep]
    \item We propose to do an in-depth investigation of the parameter-efficient tuning problem on the dense prediction tasks. To the best of our knowledge, this is the first empirical study to date that considers this problem.
    \item We design a novel Bridger that can be seamlessly integrated into any pre-trained dual-encoder vision-language models to enhance and interact with their intermediate features. It incorporates vision-specific inductive biases and task-specific information and can be integrated with prompts, adapter, and their variants.
    \item We also propose a lightweight decoder for referring image segmentation to further align visual and linguistic features.
    \item Extensive experiments and analyses demonstrate the effectiveness of the proposed approach, where it achieves comparable performance compared to existing full fine-tuning methods while updating only 1.61\% to 3.38\% parameters.\footnote{Note that we do not count the parameters of the task-specific decoder since it is required by all the baselines for the segmentation prediction.}
\end{itemize}

\section{Related work}
This work aims to design an \textit{efficient tuning} approach to \textit{referring image segmentation} built upon pre-trained \textit{vision-language models}.
In this section, we summarize previous literature and discuss the relations and differences.

\noindent\textbf{Vision-Language Models (VLMs)} target exploring a unified representation for vision and language modalities to tackle vision-and-language tasks. They can be generally divided into two types of workflow: single-stream and dual-stream.
The former includes \cite{chen2019uniter,li2020oscar,lu2019vilbert,chen2022multi,chen2022align,li2022blip}, which use a fusion module to interact the visual and textual embeddings;
The latter consists of studies, e.g., \cite{radford2021learning,jia2021scaling,li2021grounded}, which use contrastive learning to align the vision and language embeddings. Our work focus on the efficient tuning of dual-stream model due to their expansibility and the necessity of aligning features when transferring the models to downstream tasks. 

\noindent\textbf{Parameter-efficient Tuning (PET)} aims to reduce the number of trainable parameters of a pre-trained model when transferring it to the downstream tasks.
Compared with fine-tuning that retrains the whole model on a specific task, PET can make tuning a large model feasible when deploying it to each user considering the recent proliferation in model sizes.
Recent PET methods can be divided into three types:
(i) updating newly added parameters to the model or input like Adapter~\cite{houlsby2019parameter}, Prefix-tuning~\cite{li2021prefix} and Prompt tuning~\cite{zhou2022learning};
(ii) sparsely updating a small number of parameters of the model like Bit-Fit~\cite{zaken2021bitfit} and Diff Pruning~\cite{guo2020parameter};
(iii) low-rank factorization for the weights to be updated like LoRA~\cite{hu2021lora}, Compacter~\cite{karimi2021compacter} and Consolidator~\cite{haoconsolidator}. 
Adapters balance performance and extensibility in computer vision and natural language processing. Nonetheless, most current work focuses on classification and generation tasks, neglecting dense prediction tasks like segmentation and special design for multi-modal tasks. Our method addresses this gap by designing a multi-modal adapter-like module that enhances feature interaction between the two encoders of the pre-trained vision language model, which facilitates efficient transfer to downstream tasks.

\noindent\textbf{Referring Image Segmentation (RIS)} aims to segment the target objects referred by natural language descriptions by understanding the given images and expressions.
Early works can be tracked back to those CNN-LSTM-based approaches, e.g., RRN~\cite{li2018referring} and RMI~\cite{liu2017recurrent}.
They used CNN and LSTM to extract visual and linguistic features, respectively.
These features are concatenated to obtain cross-modal features, which are then fed into an FCN to perform the segmentation.
With the rapid development of Transformer, many works have begun to explore the powerful representation of the attention mechanism. Simply concatenating features from different modalities, MDETR~\cite{kamath2021mdetr} achieves great performance on different VL-tasks by feeding the fusion features into the Transformer encoder and decoder. VLT~\cite{ding2022vlt} has designed a query generation module to augment global context information, thereby enriching linguistic expressions and enhancing robustness. Taking advantage of the strong image-text alignment ability of CLIP~\cite{radford2021learning}, CRIS~\cite{wang2021cris} focus on sentence-pixel alignment to leverage multi-modal corresponding information.  
To better make use of language-related object location information for visual-linguistic interaction, PCAN~\cite{chen2022position} focuses on position-aware contrastive alignment to enhance the alignment of multi-modal features. 
Our method focuses on fusing and aligning features from different modalities using a parameter-efficient approach that leverages pre-trained vision-language models. It achieves competitive performance and scalability compared to methods using the same backbone network, while avoiding  modification of the backbone network's weight. This reduces the number of parameters to be updated and provides better hardware resource savings.

\section{Methodology}
\begin{figure*}
\begin{center}
\includegraphics[width=1.0\linewidth]{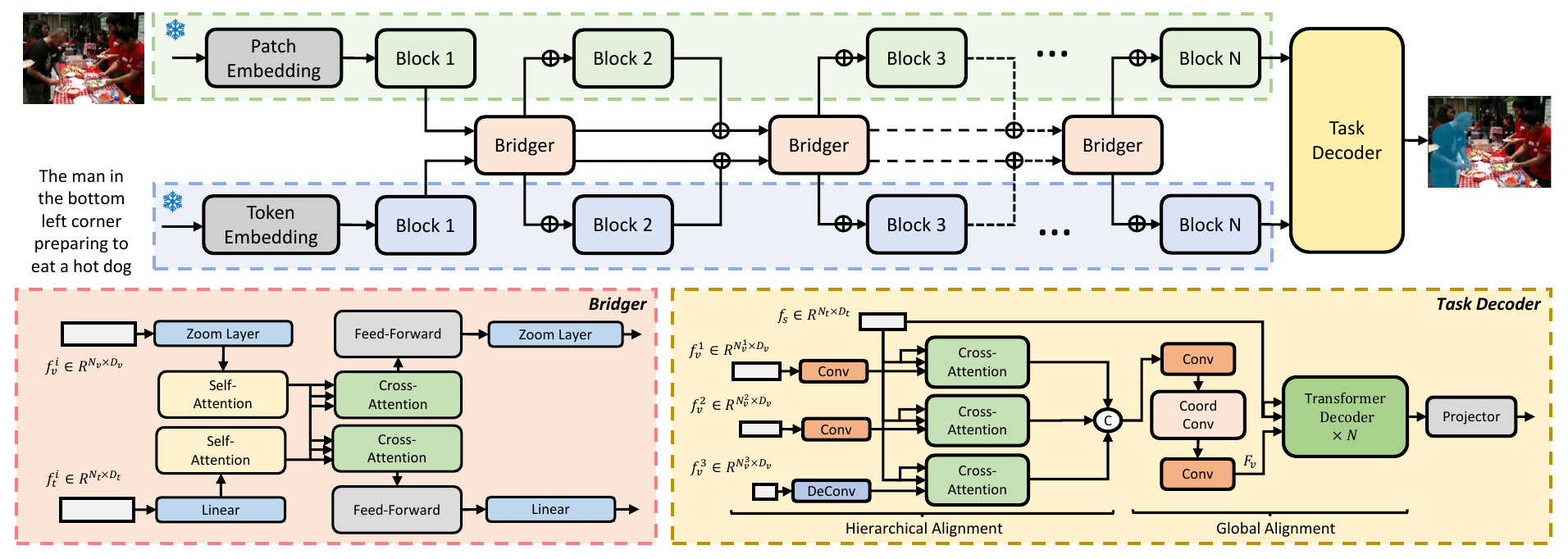}
\end{center}
   \caption{
   Given an image and a language sentence, our model extracts multiple image features $f^1_v,...,f^N_v$ from different stages of an image encoder, and word-level features $f_t$ and sentence-level features $f_s$ from a language encoder. The Bridger enables cross-modal interactions at each encoder stage. The Hierarchical Alignment Module fuses hierarchical features with global textual representations to obtain fusion features $F_v$. The Global Alignment Module combines sentence-level information with fine-grained visual features to produce the representation $F_c$. Finally, the Projector generates the mask prediction using $F_c$.
}  
\label{fig:arch}
\end{figure*}

In this section, we present the proposed parameter-efficient approach for referring image segmentation. 
As illustrated in Figure~\ref{fig:arch}, our framework contains four components, i.e., a frozen vision-language backbone (\S\ref{sec3:backbone}), a tunable Bridger (\S\ref{sec3:Bridger}), a task-specific decoder (\S\ref{sec3:dec}), and the learning objective (\S\ref{sec3:objective}). We aim to utilize the powerful pre-trained backbone as the image and text encoders for the downstream task while refraining from modifying its substantial quantity of original parameters.

\subsection{Image \& Text Feature Extraction} \label{sec3:backbone}
Given an image and a text, we extract their features through the image encoder and text encoder, respectively:
\noindent\textbf{Image Encoder.} For an input image $I\in R^{H\times W\times 3}$, we extract visual features from layers of the image encoder.
In detail, for CNN (e.g., ResNet~\cite{he2016deep}), we exploit the visual features from the last $N-1$ stages defined as $F_{v}^i, i\in \{2, ..., N\}$;
For vision Transformer~\cite{dosovitskiy2020image} (ViT), we evenly split the transformer encoders of ViT into $N$ blocks, each containing $L/N$ encoder layers. We employ the outputs of the last $N-1$ blocks to make feature interaction.
The multi-level visual features from different blocks in ViT or different stages in ResNet will be utilized in our framework as the input of the Bridger and decoder for multi-modal feature alignments.

\noindent\textbf{Text Encoder.} For an input referring expression $T$, a Transformer~\cite{vaswani2017attention} modified by \cite{radford2021learning} is used to extract text features. Similar to Image Encoder, we evenly split the transformer encoders into $N$ blocks and extract $F_{t}^i \in R^{L\times C}, i\in \{2, ..., N\}$ from different blocks of Transformer, where $C$ is the feature dimension, and $L$ is the length of the expression. 
The Transformer is applied to a lower-cased byte pair encoding representation of the text, and the text sequence is bracketed with the \texttt{[SOS]} and \texttt{[EOS]} tokens. 
The activation of the last layer of the Transformer at the \texttt{[EOS]} token is further processed to generate the global textual representation $F_s\in R^{C'}$, where $C'$ is the feature dimension.

Since the backbone image encoder and text encoder normally take the majority of the parameters, we freeze them during the tuning procedure in our approach.

\subsection{Image \& Text Feature Interaction}  \label{sec3:Bridger}
While the features from the image and text encoders do not ``see'' each other during the pre-training process, referring image segmentation requires intensive multi-modal interaction to understand the image and text features jointly.
Therefore, we propose a novel vision-and-language interaction module (i.e., Bridger) to process the intermediate features from the image and text encoders.
By doing so, the model can learn to fully use the extracted image and text features to enhance the multi-modal interaction.

Briefly, given multiple visual features $F_{v}^i, i\in \{2,...,N\}$ and linguistic features $F_{t}^i\in R^{L\times C}, i\in \{2,...,N\}$, we firstly adjust the shape of visual and linguistic features via Zoom Layer (ZL). This process can be formalized:
\begin{equation}
\begin{aligned}
& \hat{F}_v^i=\text{ZL}_{in}({F}_{v}^i) \\
& \hat{F}_t^i=\text{Linear}({F}_{t}^i)
\end{aligned}
\label{eqBridger1}
\end{equation}
where the $ZL_{in}$ means the zoom-in operation of the Zoom layer.
Afterward, we fuse the features through Interactor (ITA). This process can be mathematically expressed as
\begin{equation}
\begin{aligned}
& \hat{f}_{v}^i = \text{ITA}(\hat{f}_{v}^{i-1}+\hat{F_v^i}, \hat{f}_{t}^{i-1}+\hat{F}_t^i) \\
& \hat{f}_{t}^i = \text{ITA}(\hat{f}_{t}^{i-1}+\hat{F_t^i}, \hat{f}_{v}^{i-1}+\hat{F}_v^i)
\end{aligned}
\label{eqBridger2}
\end{equation}
Finally, we recover the dimension through the zoom layer and linear projection and make a residual connection to the original feature of the next stage (blocks) of the backbone. This process can be expressed mathematically as
\begin{equation}
\begin{aligned}
& f_{v}^i = \text{ZL}_{out}(\hat{f}_{v}^i) \\
& f_{t}^i = \text{Linear}(\hat{f}_{t}^i) \\
& F_{v}^{i+1} = F_{v}^{i+1} + f_{v}^i \\ 
& F_{t}^{i+1} = F_{t}^{i+1} + f_{t}^i \\ 
\end{aligned}
\label{eqBridger3}
\end{equation}
where the $ZL_{out}$ means the zoom-out operation of the Zoom layer.
Next, we describe their architecture in detail:

\noindent\textbf{Zoom Layer (ZL).} With the features extracted from the image and text encoders, we design a module to make dimensional changes on visual and linguistic features with consideration to the time complexity and spatial priority.
For ViT, the plain design of the architecture suffers inferior performance as a result of lacking vision-specific inductive biases. Recent studies~\cite{Xiao2021EarlyCH,he2023camouflaged} show that convolutions can help transformer to capture the local spatial contexts of images. Therefore, we reshape the feature from middle layers from ViT from $ R^{D\times C}$ to $R^{H\times W\times C}$ and use convolution to compose the Zoom layer. Moreover, for ResNet, the feature map from the first two stages can be large when the resolution of the input increases, which will make the length too long to process when using it as the input of the attention algorithm. Therefore, we adopt stride-2 2x2 convolution to reduce the size of feature maps. To unify dimensions and enlarge smaller feature maps, we use stride-2 2x2 deconvolution to enrich information. In short, when extracting the feature from the middle layers of the backbone, we use the Zoom layer to resize the feature map from the visual encoder. The process can be formalized as
\begin{equation}
\hat{F}_v^i=\left\{\begin{array}{cc}
\operatorname{Conv}\left(F_v^i\right), & h_i>=h^{\prime}, w_i>=w^{\prime} \\
\operatorname { DeConv }\left(F_v^i\right), & h_i<h^{\prime}, w_i<w^{\prime} \\
\end{array}\right.
\label{eqZoomIn}
\end{equation}
where the $h^{\prime}, w^{\prime}$ are one of the feature map's height and width from the visual backbone. 
With $\hat{F}_v^i$, we make interactive operations between feature maps of different modalities. After that, before adding the features back to the backbone, we utilize the Zoom layer to make zoom-out operations, which is the reverse process of zoom-in.

\noindent\textbf{Interactor (ITA).} With the features processed by the zoom layer, we design a module to make the interaction of modal information between the visual encoder and text encoder, which enhances the features in the middle of the pre-trained backbone while fixing the original parameters. Specifically, the Interactor is based on an attention mechanism and feed-forward network. For each feature from different modalities, we employ the original modality feature as a query and obtain the keys and values from the other modality. The interaction can be formalized as
\begin{equation}
\begin{aligned}
& \hat{f}_{v}^i=\mathcal{F}_{\text{MHSA}}(\hat{f}_{v}^{i-1}+\hat{F}_{v}^i) \\
& \hat{f}_{t}^i=\mathcal{F}_{\text{MHSA}}(\hat{f}_{t}^{i-1}+\hat{F}_{t}^i) \\
& \hat{f}_{v}^i,\hat{f}_{t}^i=\mathcal{F}_{\text{MHCA}}(\hat{f}_{v}^i, \hat{f}_{t}^i),\mathcal{F}_{\text{MHCA}}(\hat{f}_{t}^i, \hat{f}_{v}^i) \\
& \hat{f}_v^i, \hat{f}_t^i = \text{FFN}(\hat{f}_{v}^i), \text{FFN}(\hat{f}_{t}^i) \\
\end{aligned}
\label{eqIIT}
\end{equation}

\subsection{Task-specific Decoder}  \label{sec3:dec}
\noindent\textbf{Hierarchical Alignment Module.} Given multiple visual features $F^i_v, i\in \{2, ..., N\}$ from different stages and the global textual representation $F_s$, we obtain the fusion of multi-modal feature by convolution and cross-attention mechanism. For hierarchical fusion features, we simply concatenate and use a 1 × 1 convolution layer to aggregate them:
\begin{equation}
\begin{aligned}
& f_{m}^i = \operatorname{Conv}(F_v^i) \\
& f_{m}^i = \mathcal{F}_{\text{MHCA}}(f_{m}^i, F_s) \\
& F_m=\operatorname{Conv}\left(\left[f_{m}^2, ..., f_{m}^N\right]\right)
\end{aligned}
\end{equation}
where $[, ]$ is the concatenation operation, and the convolution is adopted to unify the dimension of features from different stages.
Finally, we concatenate a 2D spatial coordinate feature $F_{coord}\in R^{\frac{H}{16}\times\frac{W}{16}\times C}$ with $F_m$ and use a 3 × 3 convolution to fuse them. The visual feature $F_{v}\in R^{\frac{H}{16}\times\frac{W}{16}\times C}$ is then calculated:
\begin{equation}
F_v=\operatorname{Conv}\left(\left[F_{m}, F_{coord}\right]\right)
\end{equation}
The 2D spatial domain of $F_v$ is flattened into a sequence, forming the visual feature $F_v$, which is then used in the subsequent process.

\noindent\textbf{Global Alignment Module.} With multi-modal features gained from hierarchical alignment, we combine ample textual information corresponding to visual features by using the attention model of the Transformer. Taking the multi-modal features $F_v$ and the sentence-level feature $F_s$ as input, we firstly add the fixed sine spatial positional encoding to $F_v$ and $F_s$ respectively. Subsequently, a sequence of evolved multi-modal features $F_c$ is generated by self and cross-attention module to capture global contextual information: 
\begin{equation}
\begin{aligned}
& f_{c} = \mathcal{F}_{\text{MHSA}}(F_v) \\
& f_{c} = \mathcal{F}_{\text{MHCA}}(f_{c}, F_s) \\
& F_{c} = \text{FFN}(f_{c})
\end{aligned}
\end{equation}
where the evolved multi-modal features $F_c$ are finally utilized for the segmentation task.

\noindent\textbf{Projector.} To obtain mask prediction on each pixel according to the corresponding semantic information, we use a Projector to make transformation on cross-modal feature $F_c$ and sentence-level feature $F_s$ as Equation \ref{eqn:closs1}. 
\begin{equation}
\begin{aligned}
& F_c^{\prime}=\text{UpSample}\left(F_c\right) \\
& Z_c=\text{Conv}\left(F_c^{\prime}\right) \\
& Z_t=\text{Linear}\left(F_s\right)
\end{aligned}
\label{eqn:closs1}
\end{equation}
where $\text{UpSample}$ denotes $4\times$ upsampling, and convolution and linear projection are used to transform $F_c$ and $F_s$ into $Z_c\in R^{N\times D}, N=\frac{H}{4} \times \frac{W}{4}$ 
and $Z_t\in R^C, C=K\times K\times D + 1$. 
We split and reshape $Z_t$ into $\text{weights} \in R^{D\times K\times K}$ and $\text{bias}\in R^{D}$, where $K$ denotes the kernel size of the convolution layer. This enables it to function as a Conv2D layer, which is utilized to convert the cross-modal representation $Z_c$ into the ultimate mask prediction.

\subsection{Training Objective} \label{sec3:objective}
Considering the suboptimality of CLIP~\cite{radford2021learning}'s pre-training strategy for referring image segmentation due to its reliance on aligning the textual representation with the image-level representation, we utilize a text-to-pixel contrastive loss~\cite{wang2021cris} as our training objective, which is employed to optimize the relationship between two modalities. The objective of this contrastive loss is to ensure that $Z_t$ is similar to its corresponding $Z_c$, while being dissimilar to other irrelevant $Z_c$.
 The text-to-pixel contrastive loss can be formulated as follows: 
\begin{equation}
\begin{aligned}
L_{\text {con }}\left(Z_t, Z_c\right) & =\frac{1}{|\mathcal{P} \cup \mathcal{N}|} \sum_{i \in \mathcal{P} \cup \mathcal{N}} L_{\text {con }}^i\left(Z_t, Z_c^i\right)
\end{aligned}
\end{equation}
where $\mathcal{P}$ and $\mathcal{N}$ denote the class of 1 and 0 in the ground truth and $L_{\text {con }}^i$ is defined as: 
\begin{equation}
\begin{aligned}
L_{\text {con }}^i\left(Z_t, Z_c^i\right) & = \begin{cases}-\log \left(\sigma\left(Z_t \cdot Z_c^i\right)\right), & i \in \mathcal{P} \\
-\log \left(1-\sigma\left(Z_t \cdot Z_c^i\right)\right), & i \in \mathcal{N}\end{cases} \\
\end{aligned}
\end{equation}
where $\sigma$ is the sigmoid function. The segmentation result is obtained by reshaping $\sigma\left(Z_t\dot Z_c\right)$ into $\frac{H}{4}\times \frac{W}{4}$ and then upsampling it back to the original image size.

\begin{table*}[t]
    \centering
    \footnotesize
    \begin{tabular}{l|l|r|c|c|c|c|c|c|c|c|c}
    \hline \multirow{2}{*}{Method} & \multirow{2}{*}{Backbone} & \multirow{2}{*}{Param.~~} & \multicolumn{3}{c|}{RefCOCO} & \multicolumn{3}{c|}{RefCOCO+} & \multicolumn{3}{c}{G-Ref} \\
    \cline {4 - 12} & & & val & test A & test B & val & testA & testB & val (u) & test (u) & val (g) \\
    \hline  
    PCAN~\cite{chen2022position} & ResNet-50 & 25.56 M & 69.51 & 71.64 & 64.18 & 58.25 & 63.68 & 48.89 & $\textbf{59.98}$ & $\textbf{60.80}$ & $\textbf{57.49}$ \\
    CRIS~\cite{wang2021cris} & ResNet-50 & 40.42 M & 69.52 & 72.72 & 64.70 & $\textbf{61.39}$ & 67.10 & $\textbf{52.48}$ & $59.87$ & $60.36$ & $-$ \\
    ETRIS (Ours) & ResNet-50 & 1.68 M  & $\textbf{70.39}$ & $\textbf{73.11}$ & $\textbf{66.38}$ & 60.47 & $\textbf{67.11}$ & $50.73$ & $59.71$ & $59.95$ & $57.22$   \\
    \hline 
    RMI~\cite{liu2017recurrent} & DeepLab ResNet-101 & $61.00~\mathrm{M}$ & $45.18$ & $45.69$ & $45.57$ & $29.86$ & $30.48$ & $29.50$ & $-$ & $-$ & $-$  \\
    RRN~\cite{li2018referring} & DeepLab ResNet-101 & $61.00~\mathrm{M}$ & $55.33$ & $57.26$ & $53.95$ & $39.75$ & $42.15$ & $36.11$ & $-$ & $-$ & $36.45$  \\
    MAttNet~\cite{yu2018mattnet} & MaskRCNN ResNet-101 & $27.57~\mathrm{M}$ & $56.51$ & $62.37$ & $51.70$ & $46.67$ & $52.39$ & $40.08$ & $47.64$ & $48.61$ & $-$   \\
    CMSA~\cite{ye2019cross} & DeepLab ResNet-101 & $61.00~\mathrm{M}$ & $58.32$ & $60.61$ & $55.09$ & $43.76$ & $47.60$ & $37.89$ & $-$ & $-$ & $39.98$   \\
    CAC~\cite{chen2019referring} & DeepLab ResNet-101 & $61.00~\mathrm{M}$ & $58.90$ & $61.77$ & $53.81$ & $-$ & $-$ & $-$ & $46.37$ & $46.95$ & $44.32$  \\
    STEP\cite{chen2019see} & DeepLab ResNet-101 & $61.00~\mathrm{M}$ & $60.04$ & $63.46$ & $57.97$ & $48.19$ & $52.33$ & $40.41$ & $-$ & $-$ & $46.40$    \\
    BRINet~\cite{hu2020bi} & DeepLab ResNet-101 & $61.00~\mathrm{M}$ & $61.35$ & $63.37$ & $59.57$ & $48.57$ & $52.87$ & $42.13$ & $-$ & $-$ & $48.04$   \\
    CMPC~\cite{huang2020referring}  & DeepLab ResNet-101 & $61.00~\mathrm{M}$ & $61.36$ & $64.53$ & $59.64$ & $49.56$ & $53.44$ & $43.23$ & $-$ & $-$ & $-$ \\
    LSCM~\cite{hui2020linguistic}  & DeepLab ResNet-101 & $61.00~\mathrm{M}$ & $61.47$ & $64.99$ & $59.55$ & $49.34$ & $53.12$ & $43.50$ & $-$ & $-$ & $-$ \\
    CMPC+~\cite{liu2021cross}  & DeepLab ResNet-101 & $61.00~\mathrm{M}$ &  $62.47$ & $65.08$ & $60.82$ & $50.25$ & $54.04$ & $43.47$ & $-$ & $-$ & $49.89$ \\
    EFN~\cite{feng2021encoder}  & Wide ResNet-101 & $126.89~\mathrm{M}$ & $62.76$ & $65.69$ & $59.67$ & $51.50$ & $55.24$ & $43.01$ & $-$ & $-$ & $-$ \\
    BUSNet~\cite{yang2021bottom}  & DeepLab ResNet-101 & $61.00~\mathrm{M}$ & $63.27$ & $66.41$ & $61.39$ & $51.76$ & $56.87$ & $44.13$ & $-$ & $-$ & $-$ \\
    CGAN~\cite{luo2020cascade}  & DeepLab ResNet-101 & $61.00~\mathrm{M}$ & $64.86$ & $68.04$ & $62.07$ & $51.03$ & $55.51$ & $44.06$ & $51.01$ & $51.69$ & $-$ \\
    CRIS~\cite{wang2021cris}  & CLIP ResNet-101 & $57.31~\mathrm{M}$ & $70.47$ & $73.18$ & $66.10$ & $\textbf{62.27}$ & $68.08$ & $\textbf{53.68}$ & $59.87$ & $60.36$ & $-$ \\
    ETRIS (Ours) & CLIP ResNet-101 & $1.94~\mathrm{M}$ & $\textbf{71.06}$ & $\textbf{74.11}$ & $\textbf{66.66}$ & $62.23$ & $\textbf{68.51}$ & $52.79$ & $\textbf{60.28}$ & $\textbf{60.42}$ & $\textbf{57.86}$  \\
    \hline
    ReSTR~\cite{kim2022restr}  & ViT-B-16 & $86.19~\mathrm{M}$ & 67.22 & 69.30 & 64.45 & 55.78 & 60.44 & 48.27 & 54.48 & - & 54.48 \\
    ETRIS (Ours) & ViT-B-16 & $1.39~\mathrm{M}$ & $\textbf{70.51}$ & $\textbf{73.51}$ & $\textbf{66.63}$ & $\textbf{60.10}$ & $\textbf{66.89}$ & $\textbf{50.17}$ & $\textbf{59.82}$ & $\textbf{59.91}$ & $\textbf{57.88}$ \\
    \hline
    \end{tabular}
    \caption{Comparison with SOTA method using ResNet and ViT as backbone on the oIoU metric on RIS datasets.
    Param.: The trainable parameters of the backbone model.
    u: The UMD partition. g: The Google partition. The best results are in bold. }
    \label{table:rsota}
\end{table*}

\section{Experiments Setting}
\subsection{Datasets}
In order to assess the efficacy of each component of our method, we have conducted comprehensive experiments on three benchmarks datasets:
\begin{itemize}[leftmargin=*,noitemsep,nolistsep]
    \item \textbf{RefCOCO~\cite{kazemzadeh2014referitgame}} is a widely employed benchmark dataset for referring image segmentation. It comprises 19,994 images annotated with 142,210 referring expressions for 50,000 objects, which have been sourced from the MSCOCO~\cite{lin2014microsoft} dataset through a two-player game. The dataset is divided into four subsets, consisting of 120,624 train, 10,834 validation, 5,657 test A, and 5,095 test B samples, respectively. The average length of the expressions is 3.6 words, and each image contains a minimum of two objects.
    \item \textbf{RefCOCO+~\cite{kazemzadeh2014referitgame}} dataset consists of 141,564 referring expressions associated with 49,856 objects in 19,992 images. The dataset is divided into four subsets: 120,624 train, 10,758 validation, 5,726 test A, and 4,889 test B samples. Notably, the RefCOCO+ dataset has been constructed to be more challenging than the RefCOCO dataset by excluding certain types of absolute-location words.
    \item \textbf{G-Ref~\cite{yu2016modeling}} comprises 104,560 referring expressions associated with 54,822 objects in 26,711 images. In contrast to the two datasets described above, the expressions in G-Ref were collected from Amazon Mechanical Turk and had an average length of 8.4 words, which includes more words about locations and appearances. We present results for both the Google and UMD partitioning for G-Ref. 
\end{itemize}

\subsection{Implementation Details}
We initiate the text and image encoder with CLIP~\cite{radford2021learning}, and respectively adopt ResNet-50~\cite{he2016deep}, ResNet-101~\cite{he2016deep}, ViT-B~\cite{vaswani2017attention} as the image encoder for all ablation studies. 
We opted for CLIP because our work aims to better transfer model state from the source dataset/scenario to the target dataset/scenario via model tuning. For CLIP model, the discrepancy between model states is higher than those of GLIP and MDETR in dense prediction scenarios, which is more challenging for our Bridger design. 
We resize input images to $416 \times 416$, following the setting of CRIS~\cite{wang2021cris}. To accommodate the extra \texttt{[SOS]} and \texttt{[EOS]} tokens, RefCOCO and RefCOCO+ input sentences are limited to 17 words, while G-Ref supports up to 22 words. Our Transformer Decoder has three layers, each with 8 heads and a feed-forward hidden dimension of 512. The Projector uses a kernel size of 3 for the last convolution layer composed of $Z_t$. We train the network for 50 epochs using the Adam optimizer with a learning rate of $\lambda = 0.0001$. The learning rate of Bridger is set to $\lambda = 0.001$ for ViT and $\lambda = 0.0001$ for ResNet. We decrease the learning rate by 0.1 at the 35th epoch and train the model with a batch size of 32 on 2 NVIDIA A100 with 40 GPU VRAM.
At inference, the predicted results are upsampled to the original image size and binarized at a threshold of 0.35, producing the final result without any additional post-processing.

To evaluate the effectiveness of our model, we use two metrics commonly used in previous works: Intersection over Union (IoU) and Precision@X. IoU calculates the overlap between the predicted segmentation mask and the ground truth, while Precision@X measures the percentage of test images with an IoU score above a threshold $X\in {0.5, 0.6, 0.7, 0.8, 0.9}$. This metric assesses the model's ability to accurately localize objects.

\begin{figure*}[t]
\begin{center}
   \includegraphics[width=0.99\linewidth]{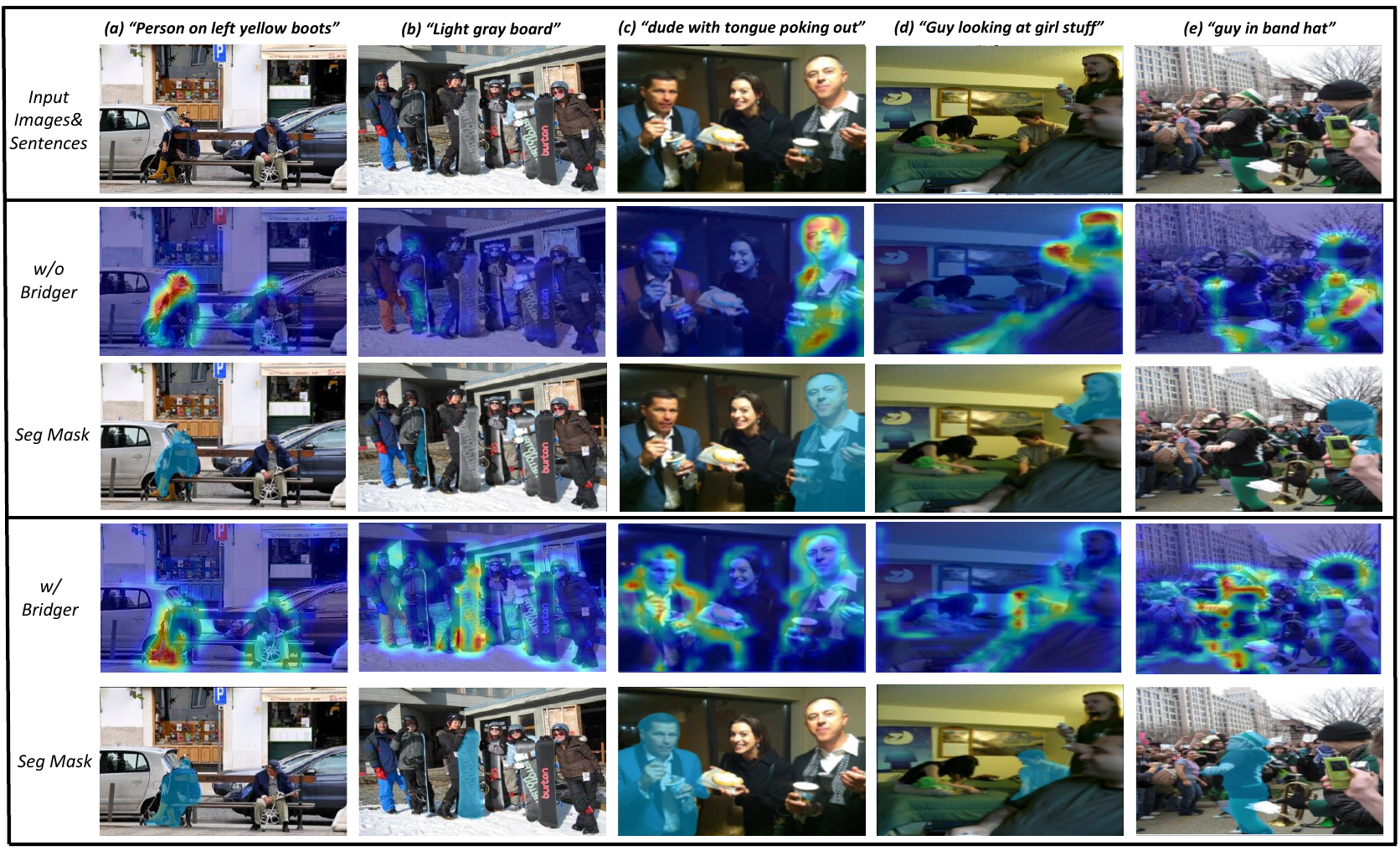}
\end{center}
   \caption{Visualization of the generated feature maps by the input images and the sentences describing objects in the images (first row). Feature maps produced by our framework $w$ Bridger (third row) contain more fine-grained features with rich edges and textures than those generated by our framework $w/o$ Bridger (second row), which is of great help for dense prediction. The fourth row demonstrates the final mask prediction of our framework $w$ Bridger. Figure best viewed in color.}
\label{fig:visualize}
\end{figure*}

\section{Experiments Results}
\subsection{Main Results}
We compare our proposed method with existing RIS methods on the same datasets, reporting the oIoU results in Table \ref{table:rsota}. To ensure a fair comparison, we categorize these methods based on their visual backbone and report their tunable parameters. Our approach achieves competitive performance on all tasks compared to existing methods using the same backbone, validating the effectiveness of our parameter-efficient approach.
Our approach's effectiveness is further amplified with increasing model scale, as observed in our experiments. This parameter-efficient approach is beneficial not only because of the strong representation abilities of pre-trained models but also due to their ability to reduce the risk of overfitting by constraining the number of parameters that require fine-tuning for downstream tasks. The Bridger plays a crucial role in early feature fusion between modalities. Additionally, our method's ability to inject vision-specific inductive biases into the pre-trained backbone reduces the performance gap between ViT-based and ResNet-based approaches. This finding also suggests the low intrinsic dimension of pre-trained models for fine-tuning~\cite{aghajanyan2020intrinsic}.

Table \ref{table:peft} compares our method with other parameter-efficient methods using oIoU metrics on RefCOCO's val-test split. To ensure a rigorous and equitable comparison, we standardized the reduction factor to 4 to minimize any potential confounding effects arising from differences in this parameter. For CoOp, we set the number of learnable tokens to 8 following~\cite{zhou2022learning}. For Conv Adapter, we applied the original method of inserting the adapter into the visual encoder. For Adapter and Compactor, we inserted adapters into both encoders. For LoRA, we incorporated LoRA into the encoders following the primary approach. Our approach achieves a 3.33\% improvement in oIoU compared to the method of freezing the weights of the backbone. Furthermore, our method shows an oIoU improvement of $1.60\% \sim 3.19\%$ over other parameter-efficient methods while using a comparable amount of fine-tuned parameters.

\begin{table}[t]
    \centering
    \resizebox{1.0\linewidth}{!}{
    \begin{tabular}{l|c|c|c|c}
    \hline \multirow{2}{*}{Method} & \multicolumn{3}{c|}{Trainable Parameters} & \multirow{2}{*}{oIoU(\%)} \\ 
    \cline { 2 - 4 } & Backbone & Prompt & Head & \\
    \hline Full-Tuning & $120.74~\mathrm{M}$ & $0.00~\mathrm{K}$ & $23.98~\mathrm{M}$ & $70.47$  \\
    Fix Backbone & $~~~~0.00~\mathrm{M}$ & $0.00~\mathrm{K}$ & $23.98~\mathrm{M}$ & $67.73$  \\
    \hline Adapter~\cite{houlsby2019parameter} & $~~~~2.39~\mathrm{M}$ & $0.00~\mathrm{K}$ & $23.98~\mathrm{M}$ & $69.46$ \\
    Conv Adapter~\cite{sung2022vl} & $~~~~1.20~\mathrm{M}$ & $0.00~\mathrm{K}$ & $23.98~\mathrm{M}$ & $69.33$ \\
    Compacter~\cite{karimi2021compacter} & $~~~~0.19~\mathrm{M}$ & $0.00~\mathrm{K}$ & $23.98~\mathrm{M}$ & $69.11$  \\
    CoOp~\cite{zhou2022learning} & $~~~~0.00~\mathrm{M}$ & $4.10~\mathrm{K}$ & $23.98~\mathrm{M}$ & $67.87$ \\
    LoRA~\cite{hu2021lora} & $~~~~0.03~\mathrm{M}$ & $0.00~\mathrm{K}$ & $23.98~\mathrm{M}$ & $68.84$ \\
    \hline ETRIS (Ours) & $~~~~1.94~\mathrm{M}$ & $0.00~\mathrm{K}$ & $23.98~\mathrm{M}$ & $\textbf{71.06}$ \\
    \hline
    \end{tabular}}
    \caption{Comparison with previous parameter efficient tuning method using Resnet101 as backbone on the oIoU(\%) metric on test-val-split of RefCOCO dataset. }
    \label{table:peft}
\end{table}

To highlight the differences between the previous and new task decoders, and their respective number of tuned parameters, we present the parameter counts in Table~\ref{table:rsota2}. These counts were calculated using their open-source code. 
For ``+'', we only calculated the parts of the model with known structures, as some modules are not open source.
Our method has fewer total tunable parameters, ranging from only 10.75\% to 20.15\% compared to other methods.

\begin{table}[ht]
    \centering
    \resizebox{1.0\linewidth}{!}{
    \scriptsize
    \begin{tabular}{l|r|r|r}
    \hline 
    \multirow{1}{*}{Method}  & \multirow{1}{*}{Tunable Param. (Backbone)} & \multirow{1}{*}{Tunable Param. (Decoder)} & \multirow{1}{*}{Tunable Param. (Total)}  \\
    \hline  
    PCAN~\cite{chen2022position}  & $25.56~\mathrm{M}$ ($6.57\%$) & $-$  & $150.21$+$~\mathrm{M}$ ($17.08\%$) \\
    CRIS~\cite{wang2021cris}  & $40.42~\mathrm{M}$ ($4.16\%$) & $42.88~\mathrm{M}$ ($50.92\%$) & $146.85~\mathrm{M}$ ($17.47\%$) \\
    ETRIS (Ours)  & $1.68~\mathrm{M}$  & $23.98~\mathrm{M}$ &  $25.66~\mathrm{M}$ \\
    \hline
    BRINet~\cite{hu2020bi}  & $61.00~\mathrm{M}$ ($3.18\%$) & $190.68~\mathrm{M}$  ($12.58\%$)& $251.68~\mathrm{M}$ ($10.30\%$)   \\
    LSCM~\cite{hui2020linguistic}   & $61.00~\mathrm{M}$ ($3.18\%$) & $80.85~\mathrm{M}$  ($29.66\%$)& $141.85~\mathrm{M}$ ($18.27\%$)  \\
    CMPC+~\cite{liu2021cross}  & $61.00~\mathrm{M}$ ($3.18\%$) & $67.66~\mathrm{M}$  ($35.44\%$)& $128.66~\mathrm{M}$ ($20.15\%$)   \\
    EFN~\cite{feng2021encoder}   & $126.89~\mathrm{M}$ ($1.53\%$) & $96.36~\mathrm{M}$  ($24.89\%$)& $232.78~\mathrm{M}$ ($11.13\%$)  \\
    CRIS~\cite{wang2021cris}   & $57.31~\mathrm{M}$ ($3.39\%$) & $40.66~\mathrm{M}$ ($58.98\%$) & $161.25~\mathrm{M}$ ($16.07\%$)  \\
    ETRIS (Ours)  & $1.94~\mathrm{M}$ & $23.98~\mathrm{M}$ & $25.92~\mathrm{M}$   \\
    \hline
    ReSTR~\cite{kim2022restr} & $86.19~\mathrm{M}$ ($1.61\%$) & $-$ & $200.63$+$~\mathrm{M}$ ($12.65\%$)  \\
    ETRIS (Ours)  & $1.39~\mathrm{M}$ & $23.98~\mathrm{M}$ & $25.37~\mathrm{M}$ \\
    \hline
    \end{tabular} }
    \caption{Comparison of parameters with existing methods, where the percentages denote the ratio of the number of tunable parameters of our method to other methods.}
    \label{table:rsota2}
\end{table}

\subsection{Qualitative Analysis}
Figure \ref{fig:visualize} demonstrates that our method with Bridger generates more detailed features with distinct edges and textures, which is superior to the model without Bridger. Bridger assists the model in better understanding the semantic information from the sentence and making more accurate positional predictions. In case (a), the feature map generated without Bridger is inadequate in accurately pinpointing the location of the boot, causing the model to focus on the person's head. Similarly, in case (b), the model fails to comprehend the location description, resulting in inaccurate predictions. Our model with Bridger generates features that capture fine-grained details and better integrate visual and textual information, as demonstrated in cases (c), (d), and (e).

\subsection{Ablation Study}
We establish the efficacy of our proposed approach by performing ablation studies on crucial modules. Further information on Bridger's hidden dimension employed can be found in Appendix~\ref{further}.

\noindent\textbf{Effect of Bridger's number and position.}
We studied the effect of the number and position of Bridger using ResNet101 as our backbone and set the scope of fusion as {[2,n], [2, n/2], [n/2, n]}, where n is the layer number of the encoder. The scope of influence of fusion features is defined as the range in which such features have an effect. For example, when fusion is performed at the first stage, the fusion features that occur at the end of the first stage will have an influence that extends to the second to nth stages of the pre-trained backbone. Table \ref{table:numpos} shows the results of different numbers of Bridgers under different scopes on RefCOCO's val-test split. 
The results indicate that Bridger scope expansion improves performance, while the number of Bridgers has little impact.

\begin{table}[h]
\centering
\begin{tabular}{l|c|c|c}
\hline Scopes & Number &  Params & oIoU(\%)  \\
\hline
$2 \rightarrow n$ & $1$ & $0.43\mathrm{M}$ & $70.96$ \\
$n/2 \rightarrow n$ & $1$ & $0.29\mathrm{M}$ & $70.18$ \\
$n \rightarrow n$ & $1$ & $1.22\mathrm{M}$ & $69.65$ \\
$2 \rightarrow n$ & $2$ & $0.72\mathrm{M}$ & $70.75$ \\
$n/2 \rightarrow n$ & $2$ & $1.51\mathrm{M}$ & $70.53$ \\
$2 \rightarrow n$ & $3$ & $1.94\mathrm{M}$ & $71.06$ \\
\hline
\end{tabular}
\caption{Ablation study of the Bridger's number and scopes.}
\label{table:numpos}
\end{table}

\noindent\textbf{Effect of ZL's component.}
We conducted experiments with various components of the Zoom Layer to analyze the optimal way for the zoom operation. Table \ref{table:ZLComponent} shows that using convolutional and deconvolutional layers for zoom-in and out operations yielded the best balance between performance and parameters. These results demonstrate that by utilizing convolution-based operations, we can adjust the size of the feature map to facilitate upcoming attention operations and augment the local information of the feature map.
\begin{table}[b]
\centering
\resizebox{1.0\linewidth}{!}{
\begin{tabular}{l|c|c|c|c|c} 
\hline
Zoom Layer ($x$) & Params & oIoU(\%)  & Pr@0.5 & Pr@0.7 & Pr@0.9 \\
\hline (a) $\text{Linear}$ & $5.77 \mathrm{~M}$ & $70.94$ & $82.89$ & $72.18$ & $\textbf{17.89}$ \\
(b) $\text{Conv}\&\text{Interpolate}$ & $1.45 \mathrm{~M}$ & $70.08$ & $81.17$ & $68.07$ & $15.95$ \\
(c) $\text{MLP}$ & $1.68 \mathrm{~M}$ & $70.62$ & $82.36$ & $71.05$ & $17.70$ \\
(d) $\text{Conv}\&\text{Deconv}$ & $1.94 \mathrm{~M}$ & $\textbf{71.06}$ & $\textbf{83.43}$ & $\textbf{72.68}$ & $17.40$ \\
\hline
\end{tabular}}
\caption{Ablation study of the Component of Zoom Layer.}
\label{table:ZLComponent}
\end{table}

\noindent\textbf{Effect of Bridger, Hierarchical Alignment Module~(HA) and Global Alignment Module~(GA).} 
We evaluated the necessity of the proposed modules by separately removing them and reporting the oIoU results on the val-test split of RefCOCO. From Table \ref{table:HG}, it can be observed that the performance decreased by 3.33\% in the absence of Bridger, 12.36\% in the absence of HA and 8.85\% in the absence of GA, which demonstrate the effectiveness of Bridger and verifies the ability of the proposed module to improve alignment.
\begin{table}[t]
\centering
\resizebox{1.0\linewidth}{!}{
\begin{tabular}{cccccc|c|c|c|c}
\hline
\multicolumn{2}{c|}{HA} &\multicolumn{2}{c|}{GA} &\multicolumn{2}{c|}{Bridger} & \multicolumn{1}{c|}{oIoU(\%)}  & Pr@0.5 & Pr@0.7 & Pr@0.9 \\
\hline
\multicolumn{2}{c|}{} & \multicolumn{2}{c|}{\checkmark}  & \multicolumn{2}{c|}{\checkmark}&\multicolumn{1}{c|}{ $58.70$ } & $69.53$ & $45.52$ & $4.33$ \\
\multicolumn{2}{c|}{\checkmark} & \multicolumn{2}{c|}{\checkmark}  & \multicolumn{2}{c|}{}&\multicolumn{1}{c|}{ $67.73$ } & $79.62$ & $68.47$ & $15.41$ \\
\multicolumn{2}{c|}{\checkmark} & \multicolumn{2}{c|}{}& \multicolumn{2}{c|}{\checkmark}& \multicolumn{1}{c|}{ $62.21$ } & $71.51$ & $53.41$ & $11.16$ \\
\multicolumn{2}{c|}{\checkmark} & \multicolumn{2}{c|}{\checkmark} & \multicolumn{2}{c|}{\checkmark} &\multicolumn{1}{c|}{ $71.06$ } & $83.43$ & $72.68$ & $17.40$ \\
\hline
\end{tabular}
}
\caption{Ablation study of Hierarchical Alignment Module Global, Alignment Module, and Bridge.}
\label{table:HG}
\end{table}

\section{Discussion}
Our method can be beneficial to other tasks such as semantic segmentation or classification due to the model's ability to facilitate early modal fusion and multi-scale feature aggregation. 
To achieve this, we propose three transformations: (1) Semantic Segmentation by considering the category name as the text, (2) Object Detection by incorporating an FPN network, and (3) Classification by making minor modifications to the decoder. More details can be seen in Appendix~\ref{further}.

\section{Conclusion}
In this paper, we propose a parameter-efficient tuning framework for referring image segmentation. In detail, for injecting vision-specific inductive biases and task-specific information into the pre-trained model while keeping its original parameters fixed, we proposed Bridger to make an interaction between the vision and language encoders. Afterward, we design a lightweight decoder to make further hierarchical and global alignment on visual and linguistic features by combining convolution and attention mechanisms. 
Our model achieves competitive performance compared to full fine-tuning on three benchmark datasets with the same backbone. Larger pre-trained models improve performance, as shown by comparisons with different visual backbones.

\section*{Acknowledgements}
This work was supported in part by the Chinese Key-Area Research and Development Program of Guangdong Province (2020B0101350001), in part by the Guangdong Basic and Applied Basic Research Foundation (NO. 2020B1515020048), in part by the National Natural Science Foundation of China (NO. 61976250), in part by the Shenzhen Science and Technology Program (NO. JCYJ20220530141211024, NO. JCYJ20220818103001002), in part by the Fundamental Research Funds for the Central Universities under Grant 22lgqb25 and in part by the Guangdong Provincial Key Laboratory of Big Data Computing, The Chinese University of Hong Kong, Shenzhen. This work was also sponsored by Tencent CCF Open Fund (NO. RBFR2022009).

{\small
\bibliographystyle{ieee_fullname}
\bibliography{iccv}}

\appendix
\clearpage
\section{Further Analysis}
\label{further}
\noindent\textbf{Effect of Bridger's hidden dim.} 
We list the oIoU results on the RefCOCO test-val set of our proposed method under different dimensions to investigate the impact of Bridger's hyper-parameters on our model's performance, as shown in Table \ref{table:hiddleDim}. The results indicate that increasing the middle dimension of Bridger leads to minor performance improvements when the dimension is less than or equal to 64. However, a slight decrease is observed when the dimension becomes 128 or larger. These findings demonstrate that our framework is robust to this hyper-parameter.

\begin{table}[h]
\centering
\resizebox{1.0\linewidth}{!}{
\begin{tabular}{l|c|c|c|c|c}
\hline Dim & Params & oIoU(\%) & Pr@0.5 &  Pr@0.7 & Pr@0.9 \\
\hline $8$ & $0.22~\mathrm{M}$ & $70.55$ & $82.56$ & $70.52$ & $16.60$ \\
$16$  &   $0.45~\mathrm{M}$ & $70.78$ & $82.98$ & $70.96$ & $17.01$ \\
$32$ & $0.92~\mathrm{M}$ & $70.92$ & $83.31$ & $71.19$ & $17.42$ \\
$64$  & $1.94~\mathrm{M}$ & $71.06$ & $83.43$ & $72.68$ & $17.40$ \\
$128$  & $4.28~\mathrm{M}$ & $70.39$ & $82.51$ & $71.12$ & $16.99$ \\
\hline
\end{tabular}
}
\caption{Ablation study of the hidden dimension of Bridger.}
\label{table:hiddleDim}
\end{table}

\noindent\textbf{Extensibility.} 
In order to further analysis the extensibility of our approach, we have integrated it with a previously established method~\cite{houlsby2019parameter,sung2022vl}, which contain more details of experiments presented in the paper.
As presented in Table~\ref{tab:add}, the integration of our approach with other methodologies yields a positive impact on the model's performance. This observation serves as evidence of the compatibility of our approach with other methodologies, and underscores the potential for effective integration to further enhance the overall performance of the model.
\begin{table}[h]
    \centering
    \resizebox{1.0\linewidth}{!}{
    \begin{tabular}{l|c|c|c|c|c|c|c|c|c}
    \hline 
    \multirow{2}{*}{Method} & \multicolumn{3}{c|}{Trainable Parameters} & \multirow{2}{*}{oIoU(\%)} & \multirow{2}{*}{Pr@0.5} & \multirow{2}{*}{Pr@0.6} & \multirow{2}{*}{Pr@0.7} & \multirow{2}{*}{Pr@0.8} & \multirow{2}{*}{Pr@0.9}  \\ 
    \cline { 2 - 4 } & Backbone & Prompt & Head &  &  &  & & &\\
    \hline Full-Tuning & $120.74~\mathrm{M}$ & $0.00~\mathrm{K}$ & $23.98~\mathrm{M}$ & $70.47$ & $82.62$ & $78.35$ & $71.35$ & $54.47$ & $17.69$ \\
    Fix Backbone & $~~~~0.00~\mathrm{M}$ & $0.00~\mathrm{K}$ & $23.98~\mathrm{M}$ & $67.73$  & $79.53$ & $74.42$ & $66.10$ & $46.39$ & $12.41$ \\
    \hline 
    Adapter~\cite{houlsby2019parameter} & $~~~~2.39~\mathrm{M}$ & $0.00~\mathrm{K}$ & $23.98~\mathrm{M}$ & $69.46$ & $81.05$ & $76.59$ & $69.69$ & $51.27$ & $16.16$  \\
    Conv Adapter~\cite{sung2022vl} & $~~~~1,20~\mathrm{M}$ & $0.00~\mathrm{K}$ & $23.98~\mathrm{M}$ & $69.33$ & $80.79$ & $76.62$ & $69.80$ & $51.69$ & $15.95$ \\    
    \hline 
    ETRIS (Ours) & $~~~~1.94~\mathrm{M}$ & $0.00~\mathrm{K}$ & $23.98~\mathrm{M}$ & $71.06$ & $83.43$ & $79.23$ & $72.68$ & $55.39$ & $17.40$ \\
    Adapter~\cite{houlsby2019parameter}+ETRIS & $~~~~4.33~\mathrm{M}$ & $0.00~\mathrm{K}$ & $23.98~\mathrm{M}$ & $71.67$ & $84.25$ & $80.12$ & $73.50$ & $56.33$ & $18.43$  \\
    Conv Adapter~\cite{sung2022vl}+ETRIS & $~~~~3.14~\mathrm{M}$ & $0.00~\mathrm{K}$ & $23.98~\mathrm{M}$ & $72.11$ & $84.46$ & $80.27$ & $73.92$ & $57.13$ & $18.74$ \\
    \hline
    \end{tabular}}
    \caption{Comparison with previous parameter efficient tuning method using Resnet101 as backbone on the oIoU(\%) metric on test-val-split of RefCOCO dataset. }
    \label{tab:add}
\end{table}

\noindent\textbf{Broader Application.}
We believe that the method could be used for other tasks such as semantic segmentation or non-dense tasks like classification:
i) The Bridge architecture facilitates early modal fusion for multi-modal tasks and multi-scale feature aggregation for dense prediction tasks.
ii) To achieve this, we propose three transformations: (1) Semantic Segmentation by considering the category name as the text, (2) Object Detection by incorporating an FPN network, and (3) Classification by making minor modifications to the decoder.
iii) In practice, for instance, our approach can achieve 88.37\% on the visual grounding task on RefCOCO when applying the Bridge to an existing multi-modal detection model (i.e., MDETR). In details, We have added bridgers that connects the visual backbone of MDETR with the text encoder, while fixing the parameters of the dual encoders. Additionally, we have incorporated an FPN (Feature Pyramid Network) to effectively merge feature maps from different stages. The fused feature are then fed forward to the decoding transformer.
In anticipation of the future, we aspire to extend the methodology by exploring its applicability to a wider range of tasks, with a particular focus on those in the vision-and-language domain.
\section{Limitation}
In this section, we conduct failure case analysis to highlight several limitations of this work. 

\begin{figure}
\begin{center}
\includegraphics[width=0.9\linewidth]{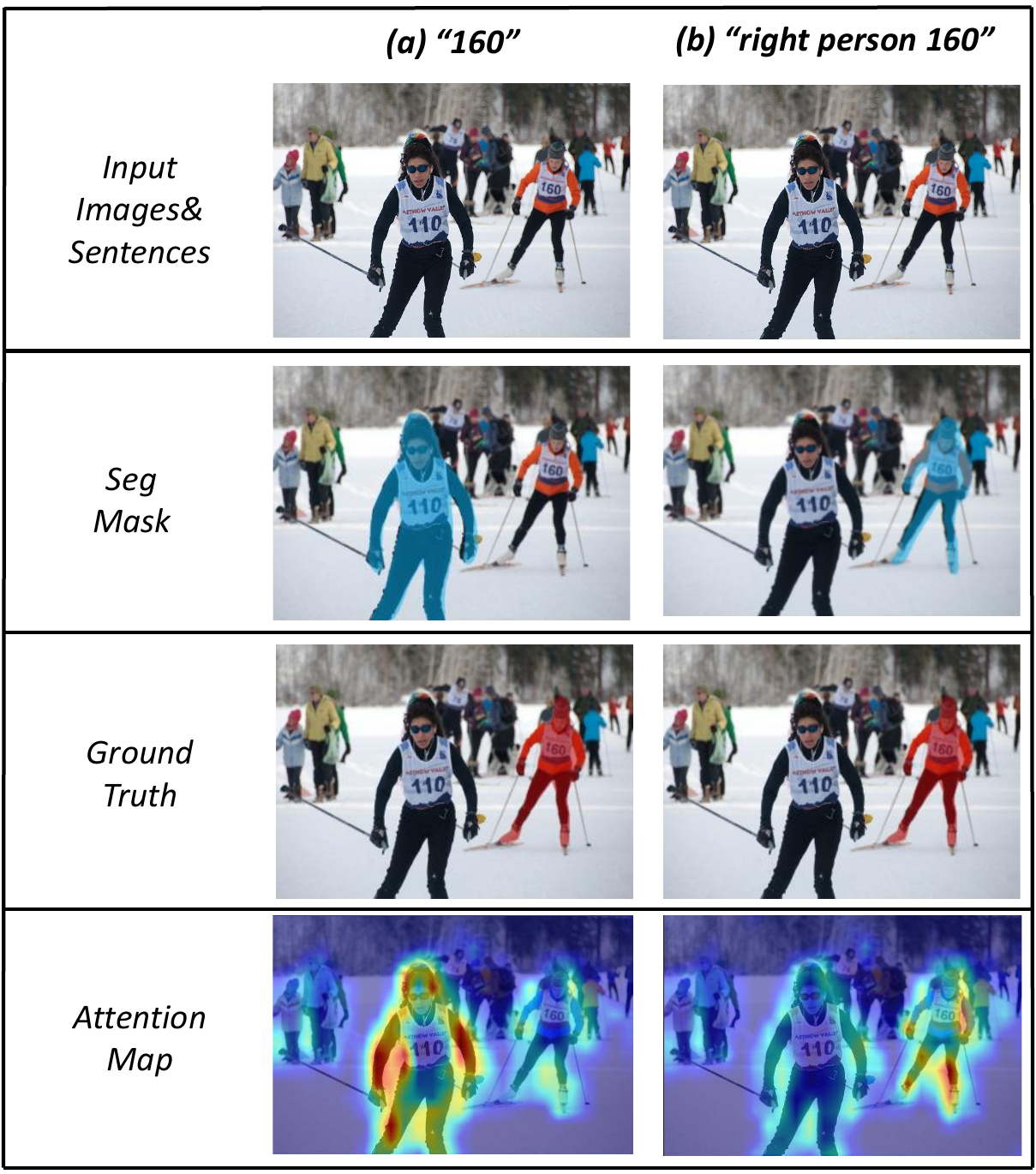}
\end{center}
\caption{Failure cases when solely describing numbers in the sentences, which show that the model may confuse visually similar numbers.}
\label{fig:case11-2}
\end{figure}

\begin{figure}
\begin{center}
\includegraphics[width=0.9\linewidth]{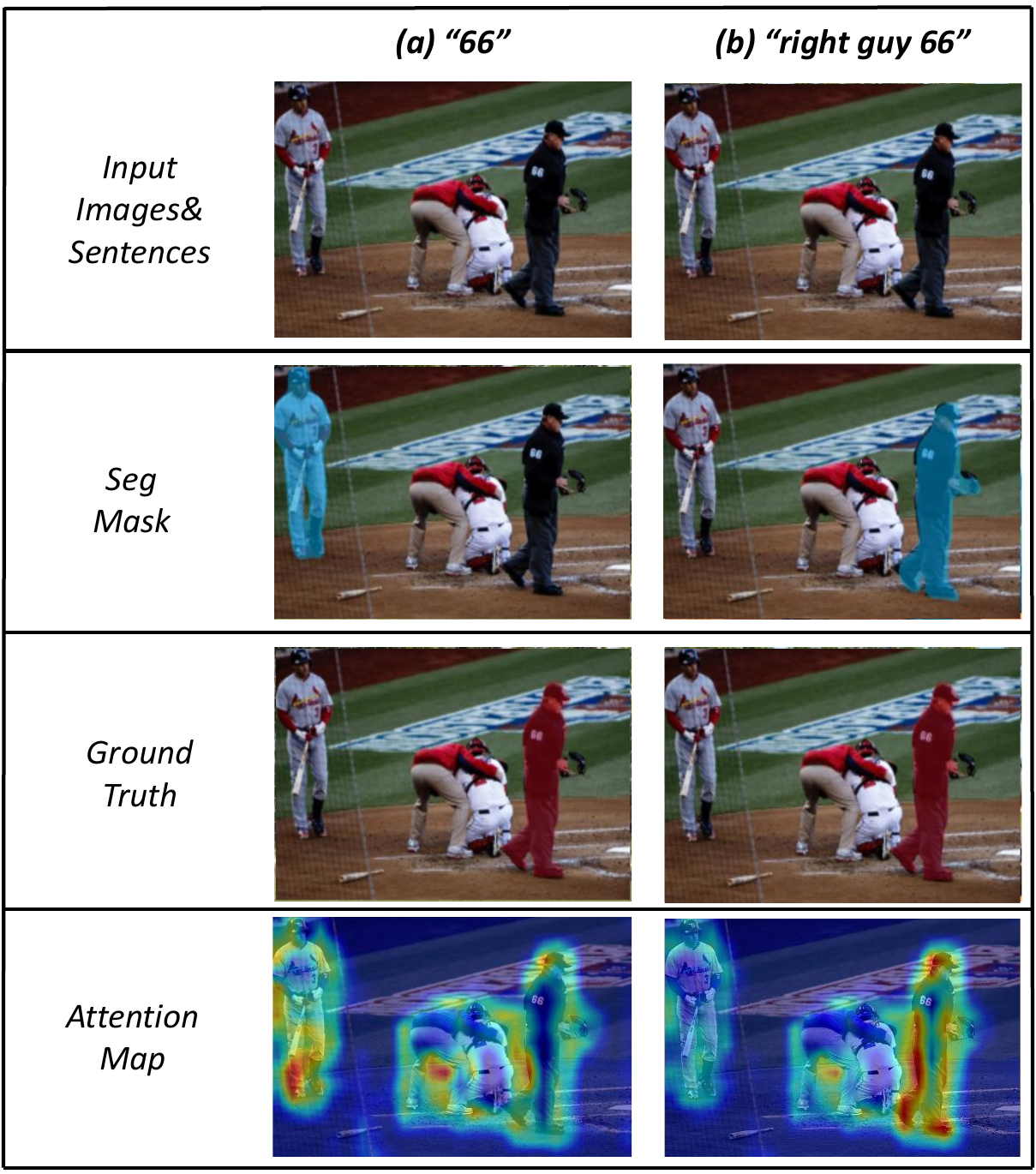}
\end{center}
\caption{Failure cases when solely describing numbers in the sentences. }
\label{fig:case11-1}
\end{figure}

\begin{figure*}[!htbp]
\begin{center}
\includegraphics[width=0.8\linewidth]{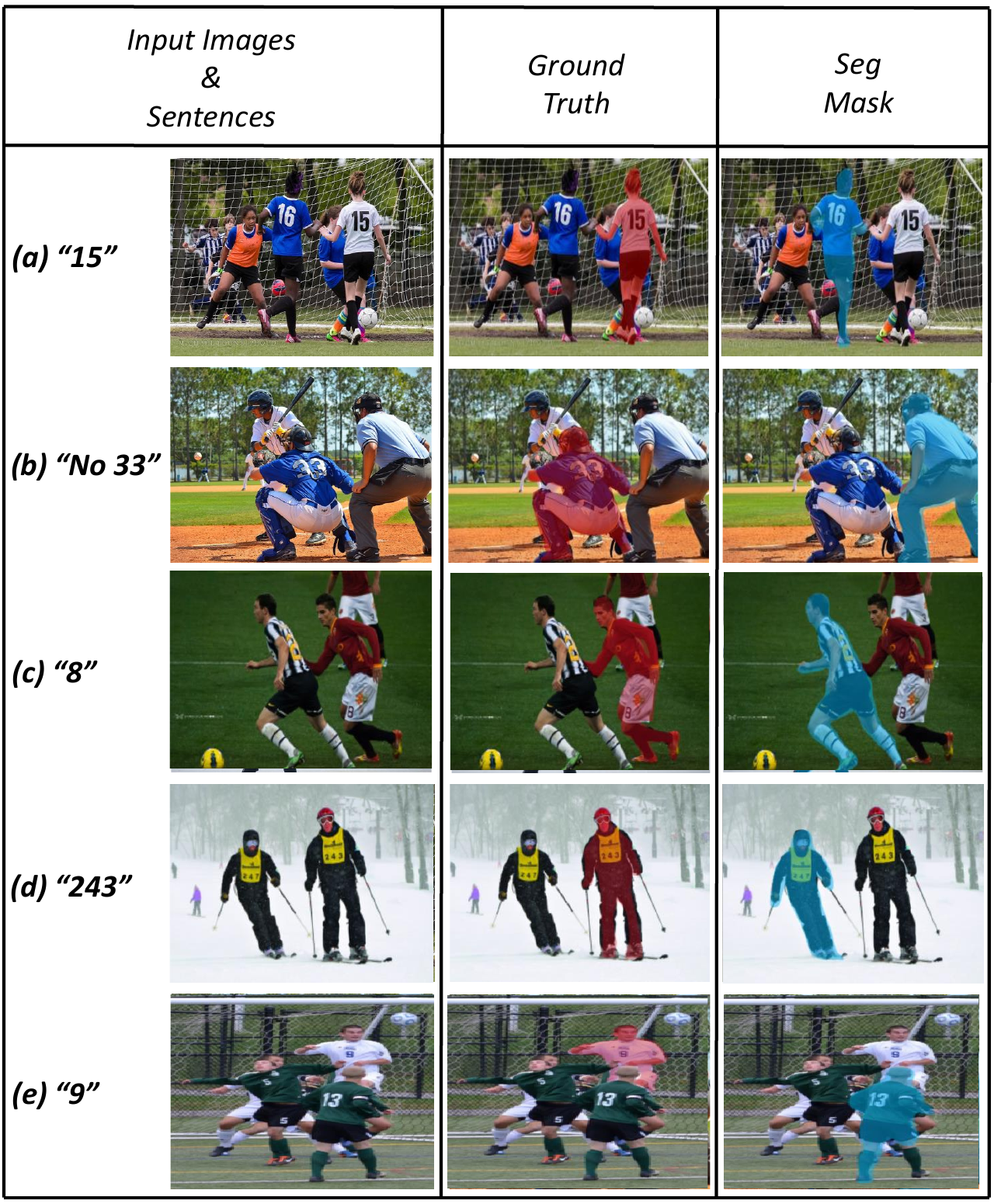}
\end{center}
\caption{Failure cases when solely describing numbers in the sentences.}
\label{fig:case12}
\end{figure*}

\noindent\textbf{Confusion on visually similar numbers.}
Figure~\ref{fig:case11-2} and Figure~\ref{fig:case11-1} presents evidence of erroneous mask predictions produced by our method, which can be attributed to a misinterpretation of digital significance within the image. The model may confuse visually similar numbers, as can be observed from the figure. However, the results also suggest that our method has a certain level of understanding of numerical meanings in both image and text contexts. More results can be seen in Figure~\ref{fig:case12}.

\begin{figure}
\begin{center}
\includegraphics[width=0.9\linewidth]{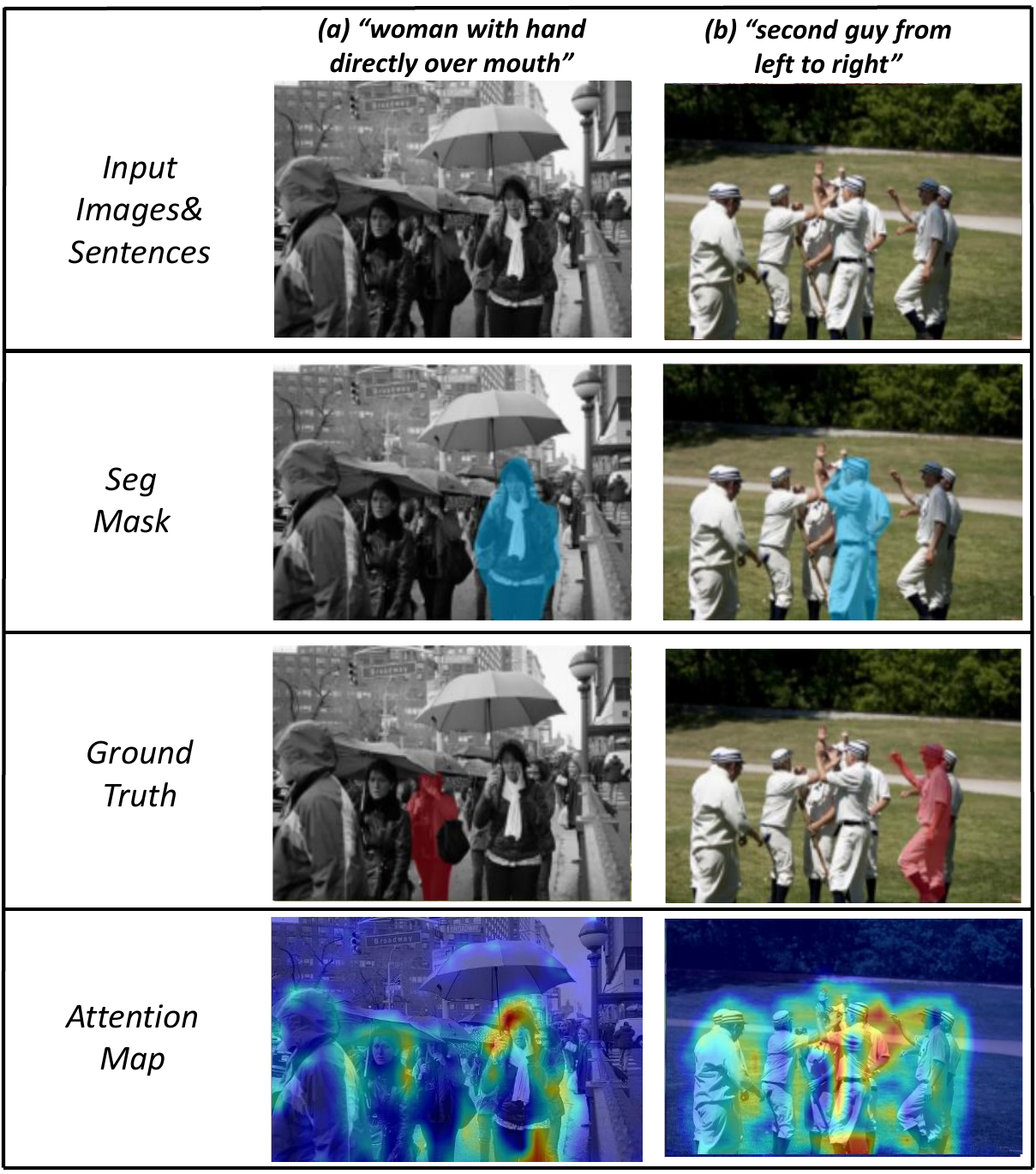}
\end{center}
\caption{Failure cases when making mask prediction for occluded objects in multi-person scenes.}  
\label{fig:case2}
\end{figure}

\begin{figure*}[!htbp]
\begin{center}
\includegraphics[width=0.8\linewidth]{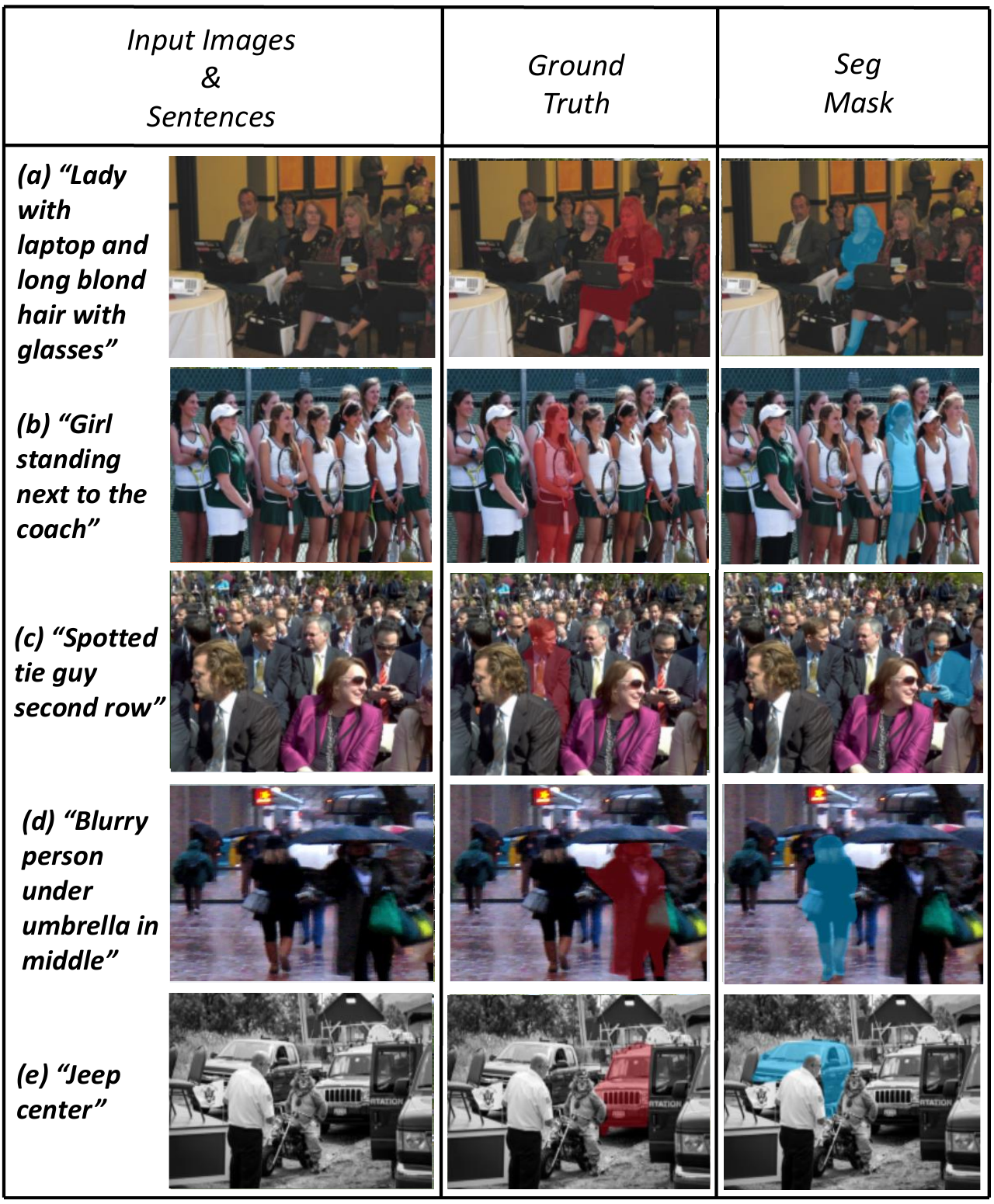}
\end{center}
\caption{Failure cases when making mask prediction for occluded objects in multi-person scenes.}
\label{fig:case22}
\end{figure*}

\noindent\textbf{Instability in processing high density of objects.}
Figure~\ref{fig:case2} shows our approach's instability of producing a precise mask in scenarios where there is a high density of individuals. In instances where the image contains a multitude of individuals, our approach may yield imprecise mask placement during the generation of corresponding masks for occluded persons.  More results can be seen in Figure~\ref{fig:case22}.

Given the aforementioned issues, future research endeavors may need to focus on augmenting the model's comprehension of linguistic information and bolstering its resilience to accurately segment occluded objects in a multi-target scene. Such efforts are crucial to improve the efficacy and reliability of computer vision systems, particularly in complex and dynamic environments.

\end{document}